\documentclass[preprint,12pt,3p]{singlecol-new}

\usepackage{natbib,stfloats}
\usepackage{mathrsfs}
\usepackage{enumitem}
\usepackage[pdftex]{graphicx}
\usepackage{epsfig}
\usepackage{resizegather}
\usepackage{xcolor}
\usepackage {amsmath}
\usepackage{caption}  
\usepackage{multirow}
\usepackage{amsfonts}
\usepackage{color}

\usepackage{footnote}
\usepackage{algorithm}
\usepackage{algorithmicx}
\usepackage{amssymb}
\usepackage{booktabs}
\usepackage{float}
\usepackage{placeins}
\usepackage{siunitx}
\usepackage{multirow}
\usepackage{multicol}
\newcommand\Tstrut{\rule{0pt}{2.4ex}}         
\newcommand\Bstrut{\rule[-0.9ex]{0pt}{0pt}}
\newcommand*{\affmark}[1][*]{\textsuperscript{#1}}

\theoremstyle{THrm}{

}

\theoremstyle{THhit}{

}

\makeatletter
\def\theequation{\arabic{equation}}

\makeatother

\begin{document}%

\setcounter{page}{1}

\LRH{Shahbano  et~al.}

\RRH{Robust Baggage Detection and Classification Based on Local Tri-directional Pattern}

\VOL{x}

\ISSUE{x}

\PUBYEAR{201X}

\BottomCatch

\CLline

\subtitle{}

\title{Robust Baggage Detection and Classification Based on Local Tri-directional Pattern}

\authorA{Shahbano\protect\affmark[1], Muhammad Abdullah\protect\affmark[1]}\affA{\affmark[1] Department of Computer Science University of Engineering and Technology Taxila, Pakistan\\
	E-mail: shahbano.ather@gmail.com, \\
	E-mail: ashar621@gmail.com, }
%
%
\authorB{Kashif Inayat\protect\affmark[2,*]}\affB{\affmark[2] Department of Electronics  Engineering, Incheon National University, South Korea \\
	E-mail: kashif.inayat@inu.ac.kr\\\affmark[*]{Corresponding author}
	}

\begin{abstract}
In recent decades, the automatic video surveillance system has gained significant importance in computer vision community. The crucial objective of surveillance is monitoring and security in public places. In the traditional Local Binary Pattern, the feature description is somehow inaccurate, and the feature size is large enough. Therefore, to overcome these shortcomings, our research proposed a detection algorithm for a human with or without carrying baggage. The Local tri-directional pattern descriptor is exhibited to extract features of different human body parts including head, trunk, and limbs. Then with the help of support vector machine (SVM), extracted features are trained and evaluated. Experimental results on INRIA and MSMT17\_V1 datasets show that LtriDP outperforms several state-of-the-art feature descriptors and validate its effectiveness.
\end{abstract}

\KEYWORD{Carrying baggage detection and classification, Local tri-directional Pattern, Support Vector Machine, Boosting Machine, Video Surveillance.}

\REF{to this paper should be made as follows: Shahbano, Abdullah, M. and Inayat, K. (2021)  `Robust Baggage Detection and Classification Based on Local Tri-directional Pattern', {\it Int. J. Internet Technology and Secured Transactions}, Vol. x, No. x, pp.xxx\textendash xxx.}

\begin{bio}
\noindent Shahbano received her MS degree in computer science from University of Engineering and Technology, Taxila, Pakistan in 2020. She did BS degree in computer science from Fatima Jinnah Women University Rawalpindi in 2017. Her research interests include Machine learning, artificial intelligence and deep learning.\vs{9}

\noindent Muhammad Abdullah received his MS degree in computer science from University of Engineering and Technology, Taxila, Pakistan in 2020. He did BS degree in computer science from Iqra university Islamabad Campus in 2016. His research interests include machine learning, deep Learning, cryptography and cyber security.\vs{8}

\noindent Kashif Inayat received his BE degree in electronics engineering from Iqra Univeristy Islamabad Campus, Pakistan in 2014, his MS degree in electronics and computer Engineering in 2019 from Hongik University, South Korea. He worked as a engineer at Digital System Design Lab at Iqra University Islamabad, from 2014 to 2017. He also worked as a researcher at R\&D Institute Incheon National University in collaboration with Samsung Electronics, South Korea. He is currently pursuing his PhD in electronics engineering at Incheon National University. His research agenda is centered on neuromorphic, embedded systems and information security. He borrow doctrine from information security, learn from machine learning and apply electronics concepts to drive his research in the following applications: AI-applications specific integrated circuits (ASIC's), tensor processing, information security, and blockchain.

\end{bio}

\maketitle

\section{Introduction}
\label{sec:1}
These days computer vision approaches have a potential impact on many intelligent surveillance systems. These approaches mainly involve knowledge, integration and management \cite{b1,b2}. Most of the detection system face some serious shortcomings like illumination conditions and complex backgrounds, and to deal with these, several approaches in the past has been proposed. In this proposed framework an effort is contributed, which attains the baggage information concerning the body of the human carrying it. Ideally it can be achieved by analysing the human in different postures while carrying bag and it deals with various texture patterns of the baggage. 
\subsection{Contributions}\label{sec:1.1}
The primary contribution of our study is 
\begin{itemize}\color{black}
	\item Handled the diverse texture patterns with the help of three dimensional link between pixels.  
	\item We proposed novel technique by combining the three dimensional link between pixels and by applying Local tri-directional Pattern descriptor to gather the information regarding the local intensities and provide accurate feature description and feature size.\color{black}
	\item Finally, the SVM classifier is used to sum up the final results and substantiate the presence of baggage in the human region.
\end{itemize}

\subsection{Organization}\label{sec:1.2}
This paper is alienated into the following segments. Section \ref{sec:2} explains the literature review of the work which had been done; Section \ref{sec:3} explains the Preprocessing Techniques. Section \ref{sec:4} explains the descriptor which is used in our framework. Section \ref{sec:5} explains the Classification strategy of SVM. Section \ref{sec:6} explains the experimental results of the proposed scheme. Section \ref{sec:7} concludes our results. 
\section{RELATED WORK}\label{sec:2}
For the detection and classification of abandoned baggage, several approaches have been proposed. Haralick et al., suggested the Gray level co-occurrence matrix (GLCM) for image classification \cite{b3,b4}. Features are extracted on the basis of co-occurrence matrix. Zhang et al. suggested a framework that uses Prewitt edge detector to extract the texture features, and isolates co-occurrence matrix for those edge images as a substitute of original images \cite{b5}. At some points in order to extract  the features, wavelet packets are used \cite{b6}. For the image retrieval and texture classification Gabor filter was also used \cite{b7}. The concatenation of Rotation invariant feature vector and Gabor filter has been used for content based image retrieval \cite{b8}. 
Wahyono et al. \cite{b9,b10} proposed a novel approach for the detection and classification of baggage and they utilize the three-dimensional information based on the human body carrying baggage. For the baggage detection Fuzzy-Model based integration framework is used and the features are then classified by using support vector machine. The results indicates that the proposed framework is one of the solutions of video surveillance system. K. Kim, et al. \cite{b11} proposed object recognition framework for different kinds of objects. Different image processing techniques and selective image methods are used to improve the image quality. The performance evaluation shows some promising results while minimizing the error rate and improving accuracy for training and testing samples.
T. Khanam et al. and Rajesh Kumar Tripathi et al. \cite{b12} - \cite{b15,b20,b23} proposes a detection and classification framework for the baggage by using the boosting strategy along with dynamic body parts. The techniques like background subtraction, HSI model, RSD-HOG features are used to deal with the challenges faced. Experimental results are better as compared to other alternatives. T. F. Ju et al. \cite{b16} proposed a vision-based object detection framework by using RADAR and LIDAR. This technique has faced several challenges like false alarm rate etc. but this paper presents an enhanced robustness in terms of vision-based object detection. W. Rakumthong et al. and Y. Tsung et al. \cite{b17,b18} proposed a framework for the boosted multiclass object detection. \textcolor{black} {Most of the aforementioned local patterns are based on the local intensity of pixels for taking the information and create a pattern according to the information provided. While the local binary pattern is based on the comparison of the neighbouring pixels and the centre pixel and assign a unique pattern to the centre pixel.} 

\textcolor{black} {In our proposed approach, there involved  three directions link among the pixels. The three directions entails the comparison of each neighbouring pixel with the centre pixel and with the two contiguous neighbouring pixels. Thus, the center-neighborhood pixels, mutual relationship of adjacent neighboring pixels is obtained, and local information based on three direction pixels are examined and gather additional information as compared to other state of the art methods. The calculated patterns involve more adjacent neighbouring pixels of each pattern value. And along with this, the magnitude pattern is also involved which is dealing with the intensity weight of each pixel. Both the magnitude pattern and Local Tri-Directional Pattern are strong enough and their concatenation provide better feature descriptor.} The results reveal that the proposed framework gives high detection rate up to 90\%.
\section{THE PROPOSED METHOD}\label{sec:3}
\subsection{An overview}\label{sec:3.1}
\begin{figure}[ht!]
	\centering{\includegraphics[width=3in]{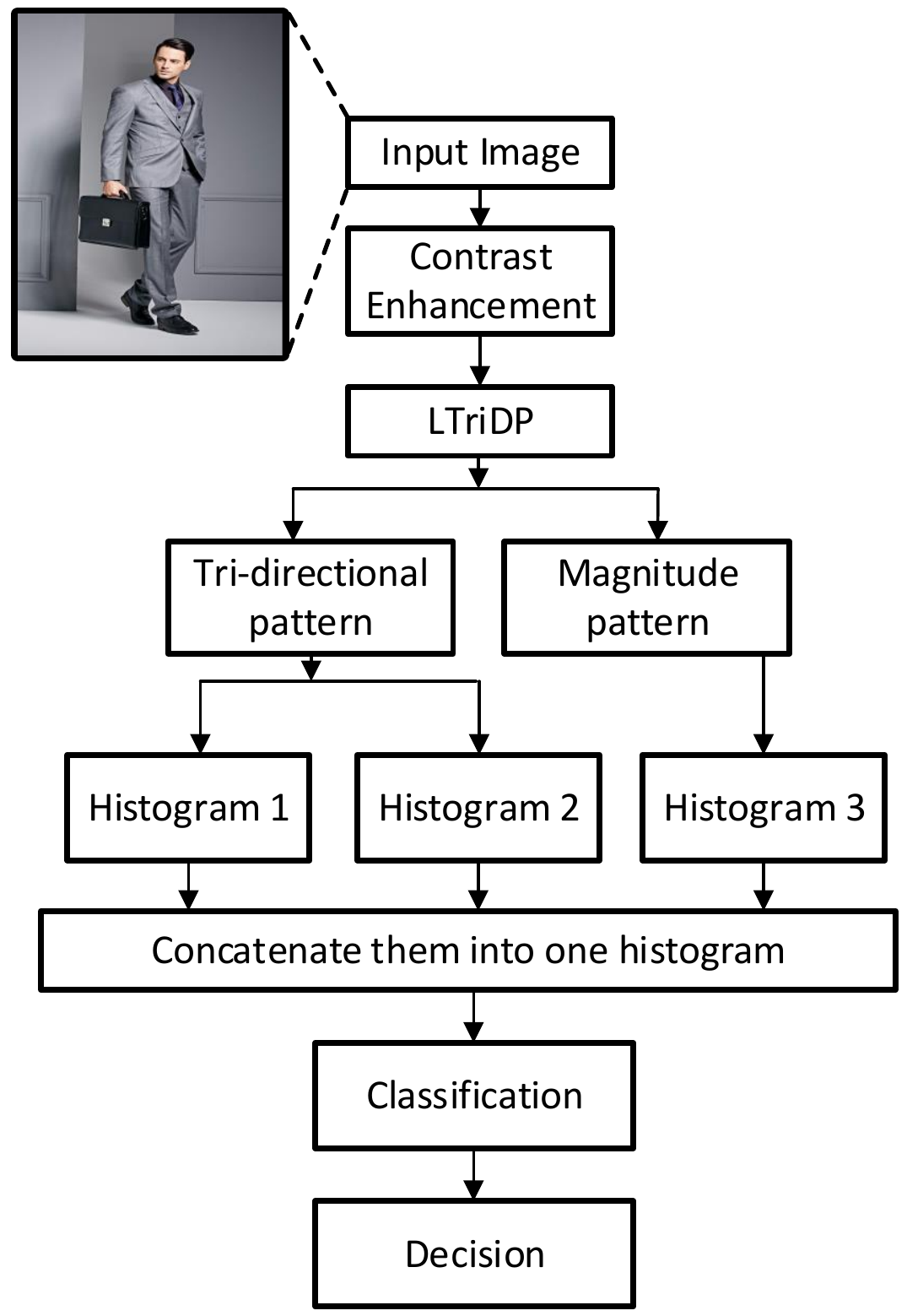}}
	\caption{\textcolor{black}{The Pipeline of the Projected Scheme}.}
	\label{fig1}
\end{figure}

This section presents the framework for the detection of human and baggage in an image which is shown in Figure \ref{fig1}. Histogram equalization technique is used in order to enhance the input images. It enhances the contrast and removes noise to increase image visual quality. Then Local tri-directional Pattern is applied on the human body part samples and baggage samples to extract features. From this, three histograms are formed, two from pattern information and one from magnitude pattern. The final decision involves concatenating the features extracted from tri-directional pattern and magnitude pattern.
\subsection{Contrast Enhancement}\label{sec:3.2}
For the contrast enhancement histogram equalization is used where the histogram of the resultant image is as flat as up to the extent. In other words, histogram equalization involves probability theory, and it is treated as the probability distribution of the gray levels. Let’s suppose $I$ to be an image whose pixel values are $I(x,y)$ . It is composed of $L$ discrete gray levels denoted by: $\{I_0, I_1,\dots,I_{L-1}\}$
. Here, $I(x,y)$ represents the intensity of the image at a spatial location $(x,y)$ with the condition that $I(x,y)\in \{I_0, I_1,\dots,I_{L-1}\}$  . The histogram of a digital image is a discrete function because intensities are all discrete values,
Histogram $h$ is defined as	

\begin{equation}\label{Eq:1}h\left(I_{k}\right)=n_k, \quad \text { for } k=0,1,2, \ldots, L-1\end{equation}

Where k-th gray level is denoted as $I_k$ and number of times gray level $I_k$ appears in the image is represented as $n_k$ . Basically, it is said that histogram is the frequency of occurrence of the gray levels in the image.
\begin{equation}f(I)=\sum_{x=0}^{I} h|x|=H|I|\label{Eq:2}\end{equation}
The intensity values in an image are considered as random values between $0$ and $(L-1)$. For normalization use Equation \ref{Eq:3}:
\begin{equation}Z_{x}=(I-Min) \frac{255}{Max-Min} \label{Eq:3}\end{equation}
Where $M i n=0, Max=(L-1)$. To find the probability of the random pixel values following formula is used [19]. \color{black}
\begin{equation}
	Pk=\frac{\text{pixels with intensity}~ k }{\text{total pixels}} \label{Eq:4}
\end{equation}
\section{FEATURE DESCRIPTOR}\label{sec:4}
\subsection{Local tri-directional Pattern Descriptor}\label{sec:4.1}
\begin{figure}[ht!]
	\centering{\includegraphics[width=3in]{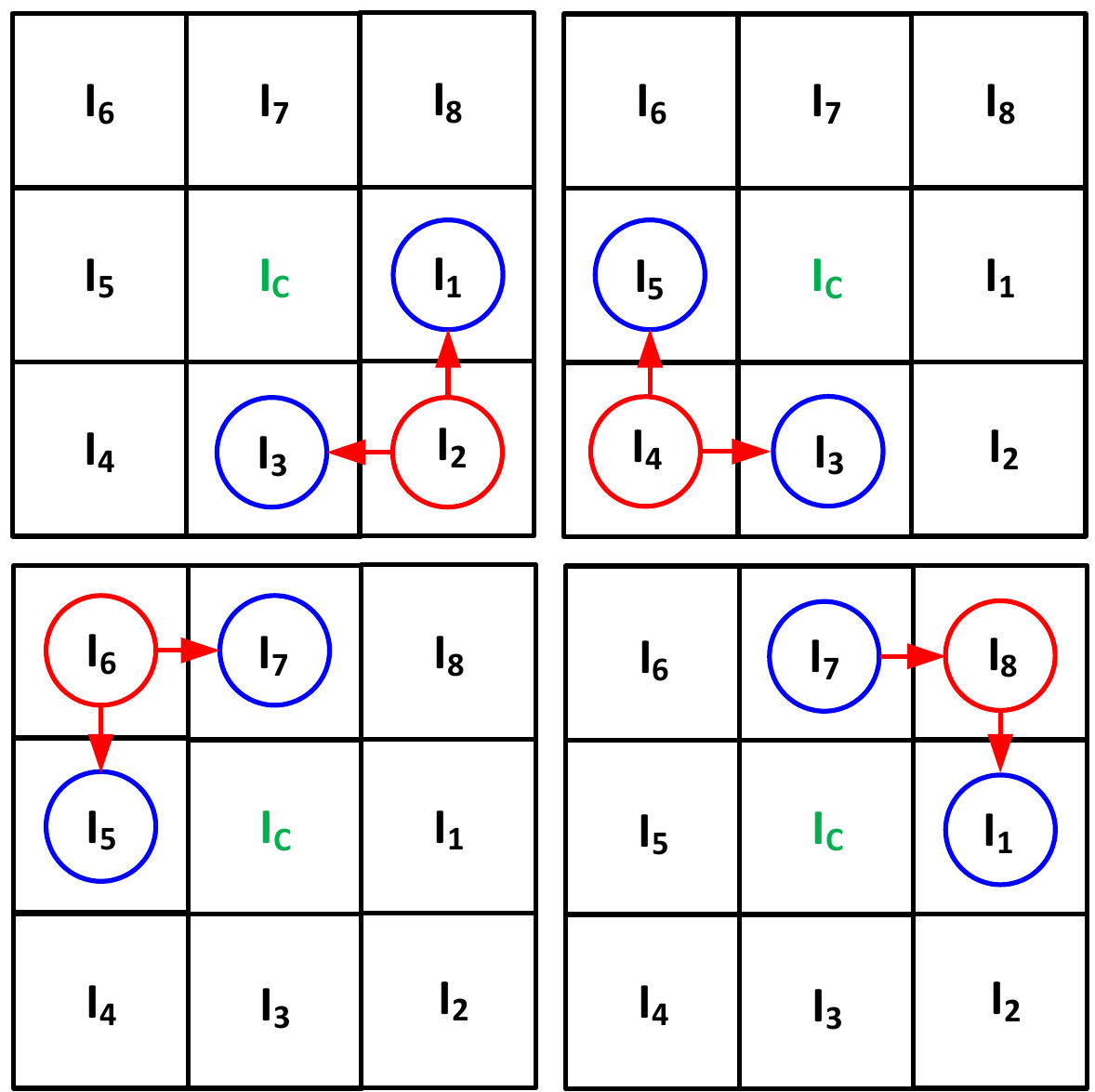}}
	\caption{\textcolor{black}{Tri-directional Patterns computation}.}
	\label{fig2}
\end{figure}

Local tri-directional pattern is an powerful variant of Local Binary Pattern (LBP) and it does not contain uniform relationship with the neighbouring pixels. It regulates in various directions while forming relationship with the neighbouring pixels.  Most of the times, 3 directions are involved in tri-directional pattern which are shown in Figure \ref{fig2}. Like the LBP, each centre pixel is surrounded by 8 neighbouring pixels and each neighbouring pixel is compared by the centre pixel and with the two contiguous neighbouring pixels. The two contiguous neighbouring pixels can be horizontal and vertical and the detailed pattern information is exhibited in Figure \ref{fig2} and explained statistically in the following equations \cite{b21,b22}. 
Lets suppose, $I_c$ as centre pixel, surrounded by 8-neighborhood pixels $I_1, I_2, I_3, I_4, I_5, I_6, I_7$, and $I_8$. Firstly, difference of each neighborhood pixel with centre pixel is calculated.
\begin{equation}
	\begin{aligned}
		D_{1}&=I_{i}-I_{8}, D_{2}=I_{i}-I_{i+1}, D_{3}=I_{i}-I_{c}\\
		\forall& i=2,3,4, \ldots, 7
	\end{aligned}\label{Eq:5}
\end{equation}

\begin{equation}
	D_{1}=I_{i}-I_{8}, D_{2}=I_{i}-I_{i+1}, D_{3}=I_{i}-I_{c}, \text { for } \mathrm{i}=1\label{Eq:6}
\end{equation}

\begin{equation}D_{1}=I_{i}-I_{i-1}, D_{2}=I_{i}-I_{1}, D_{3}=I_{i}-I_{c} \text { for } i=8\label{Eq:7}\end{equation}
Now $D_1$, $D_2$ and $D_3$ are the three differences of the neighbouring pixel with the centre pixel and further patterns are formed based on these differences. Each neighbouring pixel $i= 1,2,\dots,8$, the values of $fi(D_1,D_2,D_3)$ is calculated using Equation \ref{Eq:7} and finally the tri-directional pattern has been obtained.
To get the ternary pattern for each centre pixel, we use Equation \ref{Eq:8}, and then convert it into two binary patterns.\color{black}
\begin{equation}
	L\operatorname{Tr} i D P_{i}\left(I_{c}\right)=\left\{f_{1}, f_{2}, f_{3}, \ldots, f_{8}\right\}\label{Eq:8}
\end{equation}

\begin{equation}
	\begin{aligned}
		\operatorname{LTriDP}_{1}\left(I_{c}\right)&=\left\{S_{3}\left(f_{1}\right), S_{3}\left(f_{2}\right), S_{3}\left(f_{3}\right), \ldots . S_{3}\left(f_{8}\right)\right\}\\
		S_{3}(x)&=
		\begin{cases}
			1,& \text {when}~ x=1\\
			0, &   \text { else }
		\end{cases}
	\end{aligned}\label{Eq:9}
\end{equation}
\begin{equation}
	\begin{aligned}
		\operatorname{LTriDP}_{2}\left(I_{c}\right)&=\left\{S_{4}\left(f_{1}\right), S_{4}\left(f_{2}\right), S_{4}\left(f_{3}\right), \ldots, S_{4}\left(f_{8}\right)\right\}\\
		S_{4}(x)&=
		\begin{cases}
			1,& \text {when}~ x=2\\
			0, &   \text { else }
		\end{cases}
	\end{aligned}\label{Eq:10}
\end{equation}

\begin{equation}
	\left.\operatorname{LTriDP}\left(I_{c}\right)\right|_{i=1,2}=\sum_{i=0}^{7} 2^{l} \times \operatorname{LTriDP}_{i}\left(I_{c}\right)(l+1)\label{Eq:11}
\end{equation}

Then we get the map, and histograms are calculated by using Equation \ref{Eq:2} for both the binary patterns.
Tri-directional pattern is used to get the local valuable data, but to extract more information, more informative feature vectors are required. For this, magnitude pattern becomes very effective, and it is achieved through centre pixel, its neighbourhood pixel and two most contiguous pixels. \color{black} Following equations shows the results of magnitude patterns.

\begin{align}
	&M_{1}=\sqrt{\left(I_{i-1}-I_{c}\right)^{2}+\left(I_{i+1}-I_{c}\right)^{2}} \label{Eq:12}\\
	&M_{1}=\sqrt{\left(I_{i-1}-I_{i}\right)^{2}+\left(I_{i+1}-I_{i}\right)^{2}}~ \forall i=2,3, \ldots, 7\label{Eq:13}\\
	&M_{1}=\sqrt{\left(I_{8}-I_{c}\right)^{2}+\left(I_{i+1}-I_{c}\right)^{2}}\label{Eq:14}\\
	&M_{1}=\sqrt{\left(I_{8}-I_{c}\right)^{2}+\left(I_{i+1}-I_{c}\right)^{2}} \text { for}~i=1\label{Eq:15}
\end{align}
The values of the magnitude pattern $M_1$ and $M_2$ are calculated for each neighbouring pixel and are assigned accordingly to each neighbouring pixel.

\begin{equation}
	\begin{aligned}
		Mag_i(M_1,M_2)&=
		\begin{cases}
			1,& \text {when}~ M_1\ge M_2\\
			0, &   \text { else }
		\end{cases}
	\end{aligned}\label{Eq:16}
\end{equation}

\begin{align}
	\operatorname {LTriDP}_{\operatorname{mag}}\left(I_{c}\right)&=\left\{\operatorname{Mag}_{1}, \operatorname{Mag}_{2}, \operatorname{Mag}_{3}, \ldots, \operatorname{Mag}_{8}\right\}\label{Eq:17}\\
	\left.\operatorname{LTriDP}\left(I_{c}\right)\right|_{mag }&=\sum_{i=0}^{7} 2^{l} \times \operatorname{LTriDP}_{\operatorname{mag}}\left(I_{c}\right)\label{Eq:18}
\end{align}
At that point too, the histograms are created by using Equation \ref{Eq:2}. At the end, three feature vectors are created, two from directional pattern and one from magnitude pattern. All the three feature vectors are concatenated into one feature vector and a 150-dimensional feature vector is formed. For training and testing, we manually classified image samples as testing samples and training samples. For training purposes, the images are taken as three categories:
\begin{equation}X=\left[\text {His }\left.\right|_{L T ri D P_{1}} | \text {His}\left.\right|_{L T r i D P_{2}} | \text {His}\left.\right|_{L T r i D P_{mag}}\right]\label{Eq:19}\end{equation}

\section{CLASSIFICATION}\label{sec:5}

In the domain of image analysis, support vector machine is considered as most commonly deployed classifier.
\begin{figure}[ht!]
	\centering{\includegraphics[width=\columnwidth]{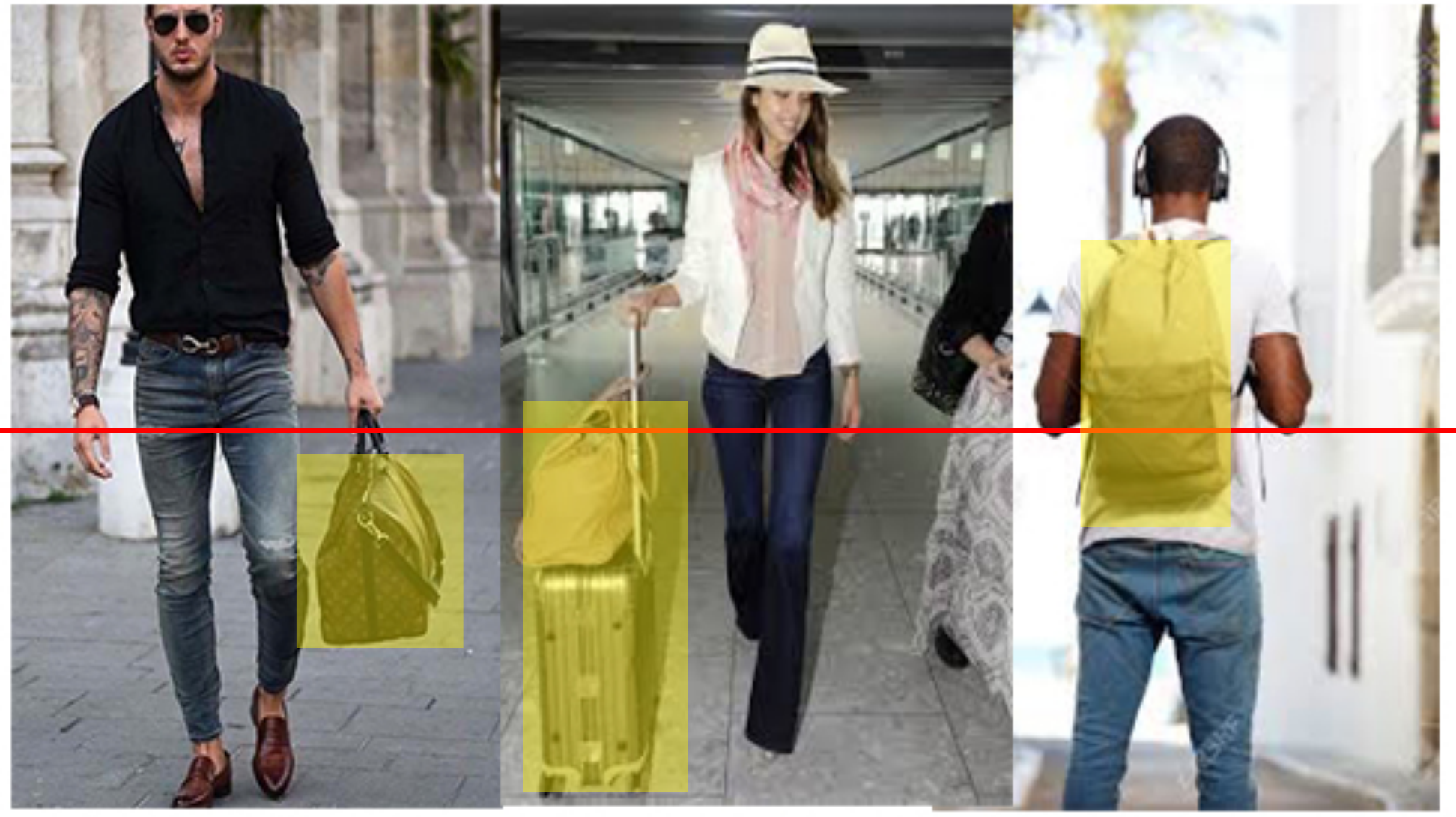}}
	\caption{\textcolor{black}{Baggage Placement Samples which are used in our work (Duffel Bag, Travel Pack or Wheeled Luggage, \textcolor{black} {Backpack})}.}\label{fig3}
\end{figure}
It finds the best distinction among different classes, as it splits the positive and negative categories with supreme space for a given training set. 
\begin{figure*}[ht!]
	\centering{\includegraphics[width=\columnwidth]{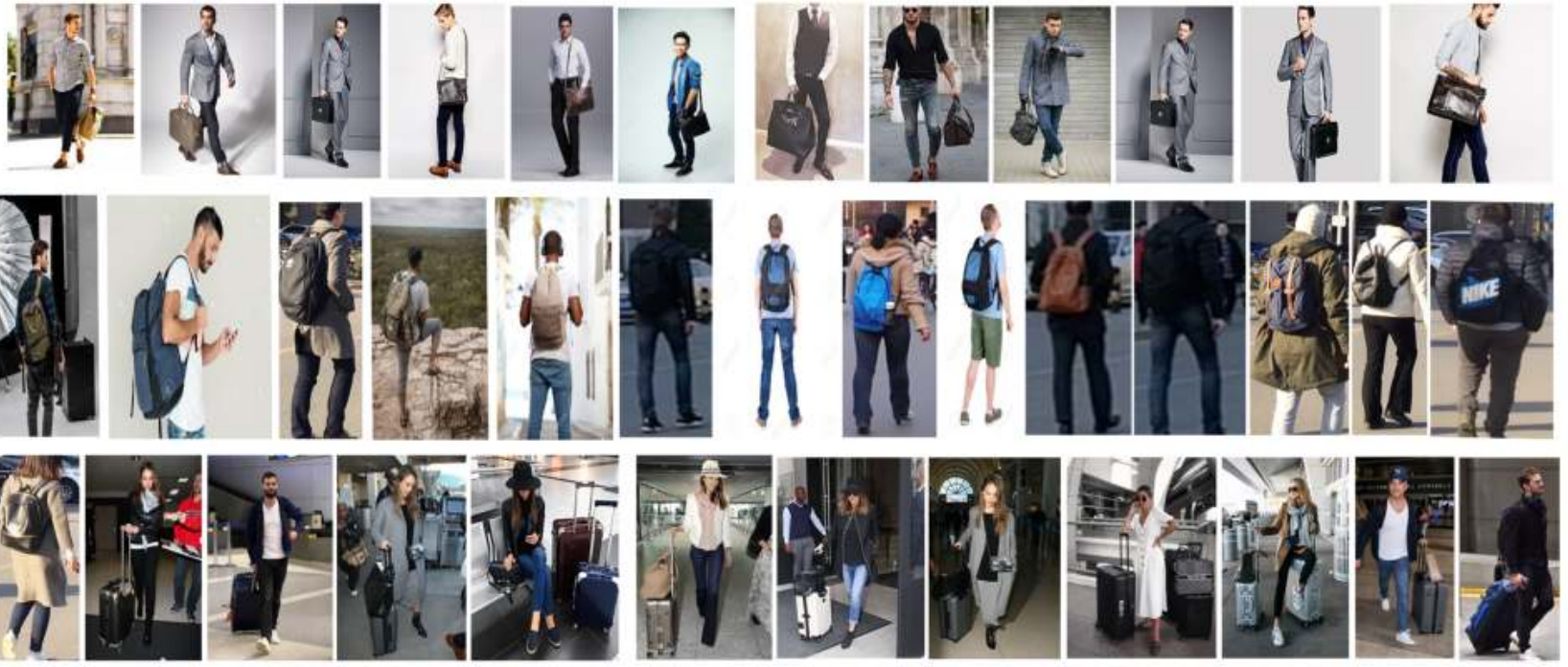}}
	\caption{\textcolor{black}{\color{black}Samples used for Training Purpose}.}
	\label{fig4}
\end{figure*}
The images selected for the training purpose are those which contains human with carrying bag or human without carrying bag. Figure \ref{fig3} shows samples of placements of baggage on the human body. (\textcolor{black} {Backpacks}, Duffel Bags, and Wheeling Bags), as shown in Figure \ref{fig4}. All these samples are resized to 256x256 and then feature descriptors are applied to them.
\section{EXPERIMENTAL RESULTS}\label{sec:6}
In this framework, techniques are implemented using Matrix Laboratory (MATLAB) which provides ability to implement the algorithm in a high-level programming language with built in Visualization Toolbox. 
\begin{table}[ht!]
	\centering
	\caption{Results of SVM Classifier with different Kernel functions.}\label{tabl1}
	\smallskip\scriptsize
	\begin{tabular}{|c|c|c|}
		\hline Validation scheme & SVM Kernel & Accuracy\Tstrut\Bstrut \\
		\hline \hline\multirow {4} {*} {$70-30$ validation } & Linear & $94.5 \%$\Tstrut\Bstrut \\
		& Quadratic & $93 \%$ \\
		& {\bf Guassian} & ${\bf95.5 \%}$ \\
		& Cubic & $95.1 \%$ \\\hline\hline
		\multirow {4} {* } { 10-Fold Cross Validation } & {\bf Linear} & ${\bf95.5 \%}$\Tstrut\Bstrut \\
		& Quadratic & $95 \%$ \\
		& Guassian & $94 \%$ \\
		& Cubic & $94.1 \%$ \\
		\hline
	\end{tabular}
\end{table}
The presented approach was first analysed to classify that the image regions having human are either with baggage or without baggage. The samples with carrying bag are considered to be as positive samples and those image samples in which human without carrying bag are negative samples. 
\begin{table}[ht!]
	\centering\scriptsize
	\caption{Evaluation of Training Dataset and comparison with other state the art methods.}\label{tabl2}
	\smallskip
	\begin{tabular}{|c|c|c|c|c|c|c|}
		\hline Category & Class. Accuracy & Precision & Recall/ TPR & Specificity & Sensitivity & FPR \Tstrut\Bstrut\\
		\hline\hline
		OUR  & $95\%$     & 0.9889 & 0.9511& 0.8333   & 0.9 & 0.8997 \Tstrut\Bstrut\\
		\cite{b23} & $90.26 \%$ & 0.9179 & 0.703 & 0.7764   &$-$& \\
		\cite{b20} & $88.27 \%$ & 0.9071 & 0.9151& 0	.8252 &     & 0.1732\\
		\cite{b11} & $81.00 \%$ & 0.8100   & 0.7700  & $-$      & $-$ &\\
		\cite{b10} & $79.86 \%$ & 0.8121 & 0.8200  & $-$      & $-$ &\\
		\hline
	\end{tabular}
\end{table}
For instance, given a training data $n$ and it belongs to 1 or -1 depends upon which class of the feature vector. It can be considered as a body with baggage 1 and body without baggage -1.  The training scheme of 70-30\% validation and 10-fold cross validation is used for the validation of results. The results after applying SVM classifier with different kernel functions on the image set is shown is Table \ref{tabl1}.\\
\begin{figure}[ht]
	\centering{\includegraphics[width=3in]{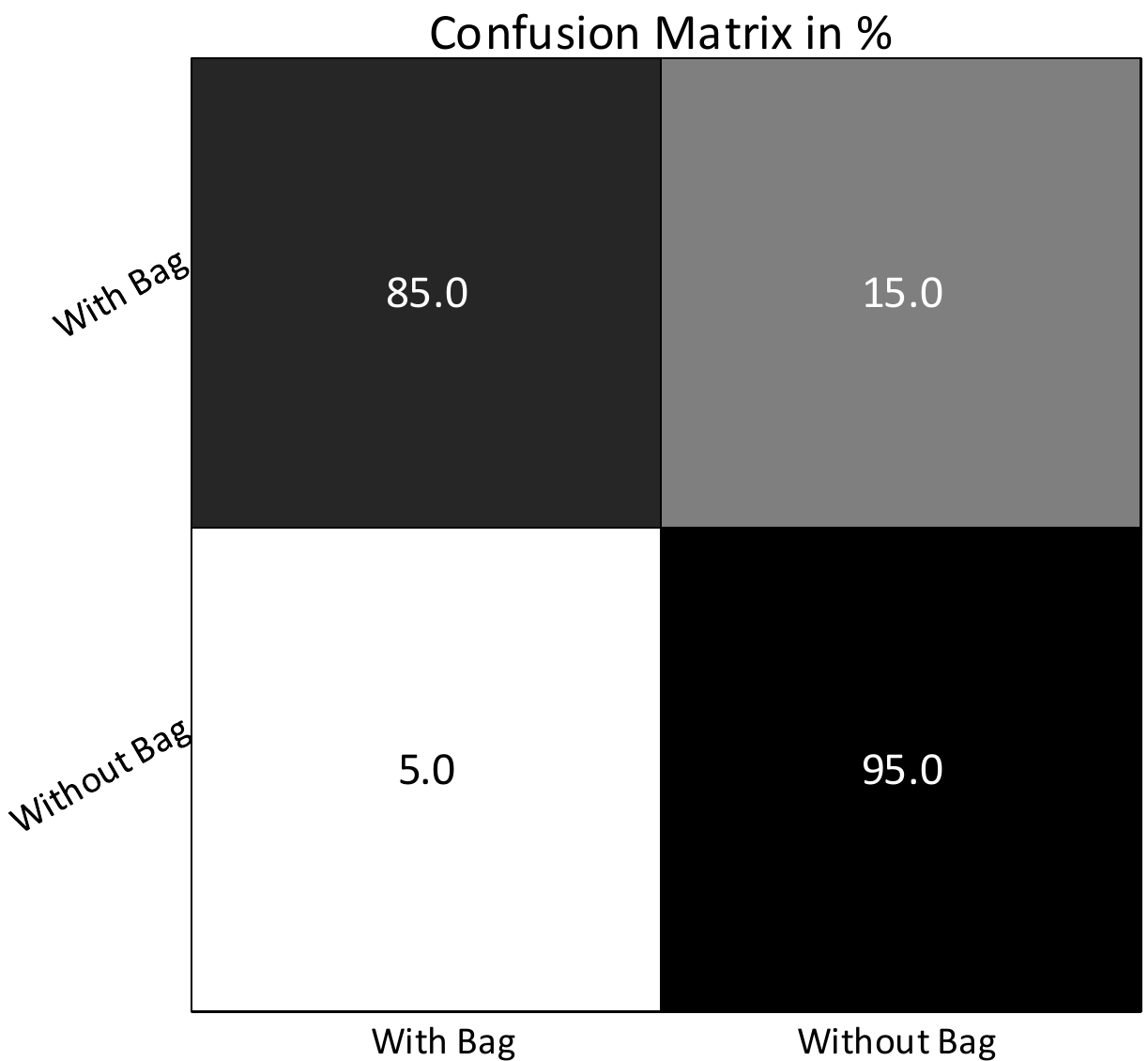}}
	\caption{\textcolor{black}{Confusion matrix demonstrating results of both classes}.}
	\label{fig5}
\end{figure}
\begin{figure}[ht]
	\centering{\includegraphics[width=3in]{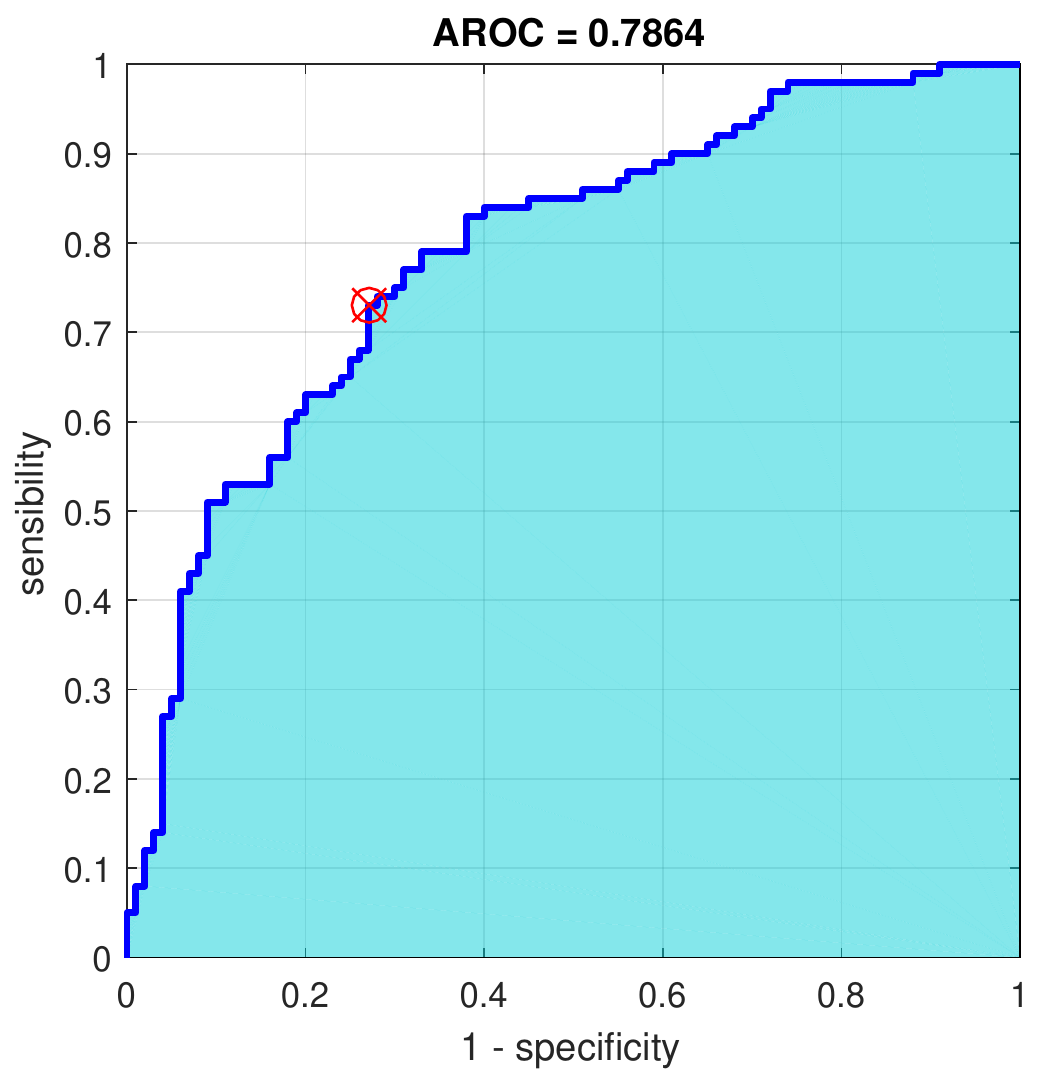}}
	\caption{\textcolor{black}{ROC Curve of the Training Dataset}.}
	\label{fig6}
\end{figure}
And Table \ref{tabl2} illustrates the assessment on the basis of accuracy and other evaluation parameters like Precision, Recall, Specificity and Sensitivity with state of the art methods. The results presented refers that the proposed method has the highest precision, recall and accuracy rates as compared to other approaches. The formulas used for computing the Precision, Recall, Specificity and Sensitivity are given below:\\
\begin{equation}Precision = TP/(TP+FP)\label{Eq:20}\end{equation}
\begin{equation}Recall = TP/(TP+FN)\label{Eq:21}\end{equation}
\begin{equation}Specificity = TN/(TN+FP)\label{Eq:22}\end{equation}
\begin{equation}Sensitivity = TP/(TP+FN)\label{Eq:23}\end{equation}
The confusion matrix and ROC Curve of the proposed framework is shown in Figure \ref{fig5} and Figure \ref{fig6}.\color{black}

\section{CONCLUSION}\label{sec:7}
In this study, an approach is proposed for the detection of human and baggage with the vision of ensuring security in public places. This approach deals with the human body parts and the baggage and utilizes a strong relationship between them. To improve the contrast of the image, histogram equalization is used. The proposed approach attains the three-dimensional information of the baggage from the body of the human carrying it. On the extracted regions Local tri-directional Pattern is applied and get feature vectors. Finally, the extracted feature vectors of pattern information and magnitude information are concatenated into one feature vector and trained by the Boosting Support Vector Machine (SVM). After conducting extensive experiments, the proposed system shows a satisfactory classification accuracy rate of 95\%. One of the limitations of this framework is that, it fails to detect baggage in much occluded   state where objects are not visually separable. Basically, in such occluded state it is not possible to detect the edges. However, potential promotion of these limitations will improve the system performance in a very notable time.




\end{document}